# Dense Object Detection Based on De-homogenized Queries

**Yueming Huang [1], Chenrui Ma [2], Hao Zhou [1], Hao Wu [1] and Guowu Yuan [1,*]**

[1] School of Information Science and Engineering, Yunnan University, Kunming 650504, China; huangyueming@mail.ynu.edu.cn (Y.H.); zhouhao@ynu.edu.cn (H.Z.); haowu_sise@ynu.edu.cn (H.W.)

[2] School of Computer Science and Engineering, Central South University, Changsha 410083, China; 8208210320@csu.edu.cn

* Correspondence: gwyuan@ynu.edu.cn

**Abstract:** Dense object detection is widely used in automatic driving, video surveillance, and other fields. This paper focuses on the challenging task of dense object detection. Currently, detection methods based on greedy algorithms, such as non-maximum suppression (NMS), often produce many repetitive predictions or missed detections in dense scenarios, which is a common problem faced by NMS-based algorithms. Through the end-to-end DETR (DEtection TRansformer), as a type of detector that can incorporate the post-processing de-duplication capability of NMS, etc., into the network, we found that homogeneous queries in the query-based detector lead to a reduction in the de-duplication capability of the network and the learning efficiency of the encoder, resulting in duplicate prediction and missed detection problems. To solve this problem, we propose learnable differentiated encoding to de-homogenize the queries, and at the same time, queries can communicate with each other via differentiated encoding information, replacing the previous self-attention among the queries. In addition, we used joint loss on the output of the encoder that considered both location and confidence prediction to give a higher-quality initialization for queries. Without cumbersome decoder stacking and guaranteeing accuracy, our proposed end-to-end detection framework was more concise and reduced the number of parameters by about 8% compared to deformable DETR. Our method achieved excellent results on the challenging CrowdHuman dataset with 93.6% average precision (AP), 39.2% MR$^{-2}$, and 84.3% JI. The performance overperformed previous SOTA methods, such as Iter-E2EDet (Progressive End-to-End Object Detection) and MIP (One proposal, Multiple predictions). In addition, our method is more robust in various scenarios with different densities.

**Keywords:** object detection; dense detection; DETR; transformer

## 1. Introduction

Dense object detection is an essential task in computer vision, aiming at accurately detecting and locating multiple mutually occluded objects from an image or video. There is a trade-off in the design of dense object detection algorithms. On the one hand, the detector has to integrate the coded information and regress each target as much as possible to avoid miss detection. On the other hand, while pursuing a higher recall, it is important to prevent causing duplicate predictions, i.e., multiple predictions corresponding to the same GT (Ground Truth). This trade-off is fundamental in dense detection tasks.

Among the current detection methods, there are two main ways to solve the problem of missed detection and repeated prediction as follows: (1) box-level post-processing de-duplication strategy; (2) matching the strategy of candidate boxes to the set of GTs during the training process, i.e., to determine which candidate anchor or query should predict which GT, and which one should not. They correspond to detection algorithms based on anchors and post-processing and end-to-end detection algorithms based on queries.

In the widely used anchor-based detection algorithms [1–4], training is conducted using manually set dense anchors for many-to-one matching with GTs. Each anchor predicts the nearest IoU distance to itself in the neighborhood, as shown in Figure 1a. The many-to-one matching training strategy and the absence of additional penalties for repeated predictions in these methods make the network inherently incapable of de-duplication. This results in multiple neighboring anchors regressing to the same GT during the prediction, so additional post-processing methods, such as non-maximum suppression (NMS), are required to remove duplicate predictions. However, this greedy algorithm-based box-level method, NMS, which sets a fixed IoU threshold for removing highly overlapping predictions and adjusts IoU suppression thresholds to balance the contradiction between repeated prediction and missed detection, is still unable to solve the contradiction in different dense scenarios. In the subsequent improvement algorithms, SoftNMS [5] mitigates this contradiction by setting the soft threshold for box suppression, and adaptive NMS [6] adaptively changes the threshold for suppression. However, box-level NMS and other greedy-based post-processing methods still perform poorly in dense scenes.

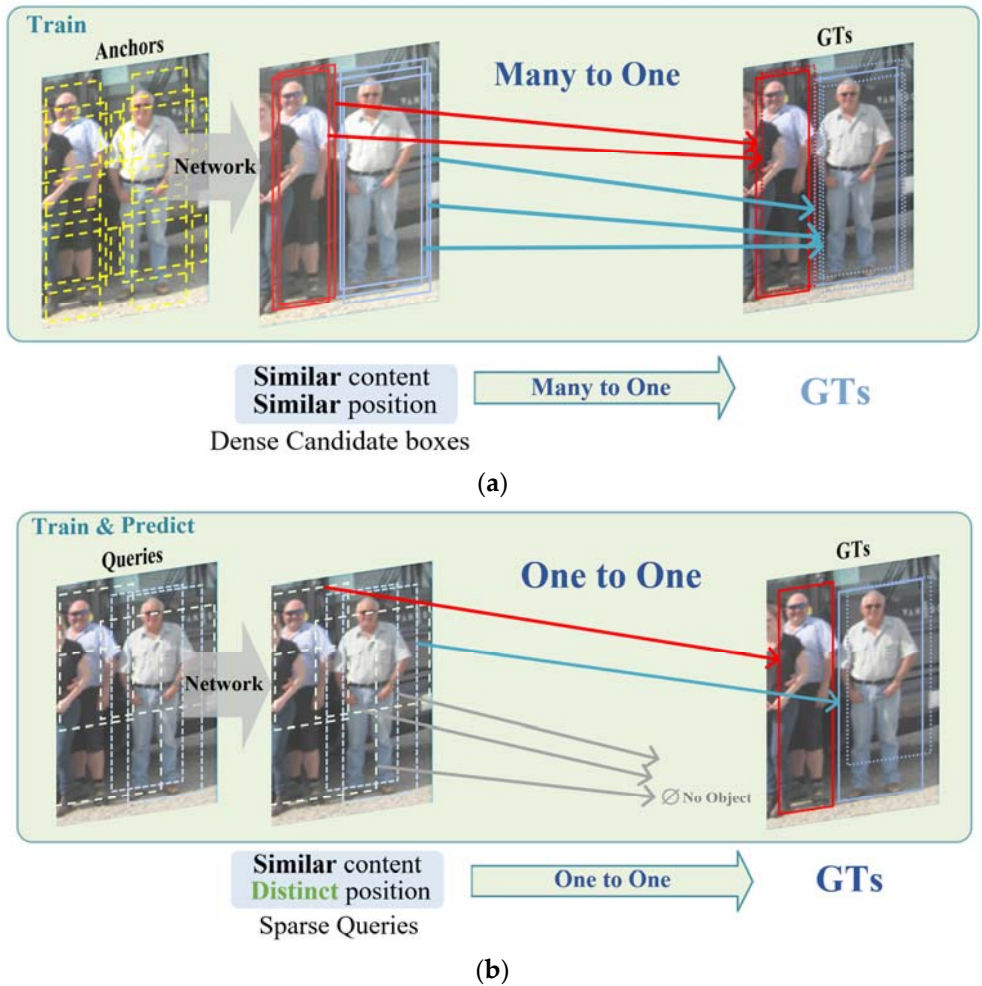

(**a**)

(**b**)

**Figure 1.** Current mainstream detection frameworks, different coloured arrows are matched with the same coloured GTs. (**a**) Anchor-based detector; (**b**) query-based detector.

In the end-to-end query-based detection algorithm [7,8], as shown in Figure 1b, a global one-to-one matching algorithm based on bipartite graph matching is used to replace the previous many-to-one greedy matching algorithms in training, and a learnable sparse query is used to replace the manually set dense anchors. During the training process, each query learns to focus on objects of different regions and shapes according to statistical laws and performs global scale segmentation matching on the set of GTs (the queries that should predict which GTs should and should not be used). During training, the network learns the one-to-one matching of queries for GTs, and there is a loss penalty for repeated predictions during training, so the model learns the de-duplication capability inside the network through query location segmentation without additional post-processing de-duplication such as NMS.

However, in query-based detection algorithms, queries with similar locations focus on similar encoding information, and queries with similar content tend to make repeated predictions after the same regression and classification headers, which means that such a problem is worse in dense scenes. In addition, since similar queries do not have enough differentiated information for the network to learn the ability of duplicate removal in training, similar queries bring unstable backpropagation during the training, which reduces the learning efficiency of the encoder in crowded scenarios.

*1.1. Solving the Query Homogenization Problem*

Query homogeneity is mainly the result of two reasons: content homogeneity and location homogeneity. For location homogenization, DN-DETR, and DINO-DETR [9,10] improve the convergence speed by limiting the distance of query fine-tuning in the training phase so that the matching division between GT and the query becomes more stable. In the work of DDQ [11], good results were achieved by adding non-maximal suppression (NMS) to the initialized query process to filter the initialized positions and suppress the queries with similar positions. However, due to the high density of GTs in dense scenarios, the location similarity is very high, so the differentiation of query locations can only alleviate the repetitive prediction in dense scenarios. Secondly, the computational cost of non-maximal suppression (NMS) is too high, and the experiments in RT-DETR [12] demonstrate the high time cost of NMS.

In this paper, we address this problem by differentiating the content of the query. When query locations are close, the encoded content is homogeneous due to similar attentional scopes. Therefore, we designed a learnable differentiated encoding for each query to break the strong correlation between the content of the query and its location. In this way, even if the predicted locations of the queries are close to each other, the model can adopt different learning strategies with differentiated information. This can significantly improve the de-duplication ability of the decoder and the encoder. As shown in Figure 2, the colors of the query prediction boxes represent their content differences; we aim to propose a detection framework based on content differentiation. Meanwhile, in generating differential encoding, the query can exchange differential information with the surrounding prediction, replacing the self-attention among the queries in the original decoder and simplifying the network structure.

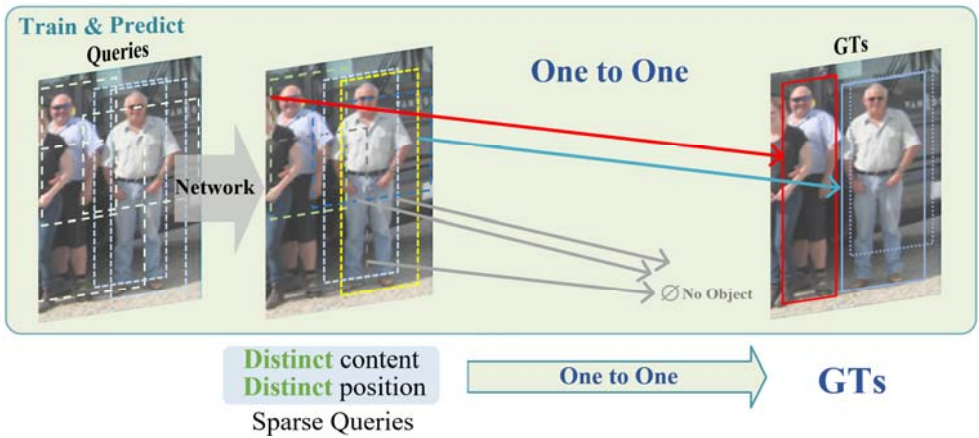

**Figure 2.** Our detection framework based on differentiated query. Different colored arrows are matched with the same colored GTs, and different colored boxes represent differences in the encoded content of the queries.

## 1.2. Alligned Decoder

In DETR-like detectors, the encoder generates feature maps for the decoder to query, allowing for layer-by-layer fine-tuning to achieve more precise predictions. However, the structural differences between the encoder and decoder lead to disparities in their encoding and decoding methodologies. Much work has been conducted to improve the end-to-end detector by aligning the design of encoders and decoders, enabling the more effective integration of information during cross-attention and enhancing model efficiency. For example, DAB-DETR [13] and conditional DETR [14] add the same positional coding in the decoder as in the encoder to align the coding and decoding methods, which reduces the information discrepancy caused by the difference between the encoder and the decoder methods and obtains an improvement in accuracy. Inspired by these papers, we designed the decoder as a structure aligned with the encoder, eliminating the self-attention among the queries in the decoder. At the same time, using the asymmetric difference aggregation (ADA) mechanism in the difference encoding process can be a good substitute for the communication between queries. Without decreasing the detection efficiency and accuracy of the model, the method in this paper can reduce the number of parameters by about 8%.

## 1.3. Query Initialization Considering Both Position and Confidence

In a two-stage DETR-like detection model, the query is usually initialized using the encoder's output confidence score Top-K algorithm. Due to the separate setting of confidence and location loss functions, there is a mismatch between the confidence score and IoU when selecting the initialized query, i.e., there are candidate predictions with high IoU but low confidence, resulting in them being filtered out by the Top-k algorithm. For this reason, this paper proposes using the joint loss of confidence and GIoU to supervise the training of the encoder output and to optimize the mismatch between the confidence and IoU to initialize the query more efficiently and achieve better detection results with fewer queries.

### 1.4. Our Contribution

In this paper, we address the problem of homogeneous queries in query-based detectors by proposing learnable differential encoding to add to the query, which increases the de-duplication capability in the network while improving the learning efficiency of the encoder and decoder; secondly, we used a higher-quality initialization of the query by taking into account both the positional and confidence losses; and finally, we optimized the structure of the decoder to reduce the number of model parameters without significantly affecting the model accuracy. With the proposed differential query learning strategy, the method in this paper outperformed the recent SOTA methods of Iter-E2EDet (Progressive End-to-End Object Detection) [15], MIP (One proposal, Multiple predictions) [16], and AD-DETR (Asymmetrical Decoupled Detection Transformer) [17], while the parameters were reduced by about 8% with fewer decoders.

## 2. Analysis of Similar and Differentiated Queries

This section mainly discusses how similar queries in query-based detectors can lead to repeated predictions and training inefficiency in dense scenarios. Queries with similar locations focus on similar encoded feature information, which results in content-similar queries, and they are more inclined to make similar duplicate predictions when passing through the same fully connected classification and regression heads.

In Figure 3, we statistically represent the query content's similarity and relative IoU distance in deformable-DETR. The vertical coordinate is the cosine similarity between the queries, and the horizontal coordinate is the IoU distance between the queries. The encoded content and location of the queries are strongly correlated. The closer the IoU distance between queries, the greater the probability that they have high content similarity.

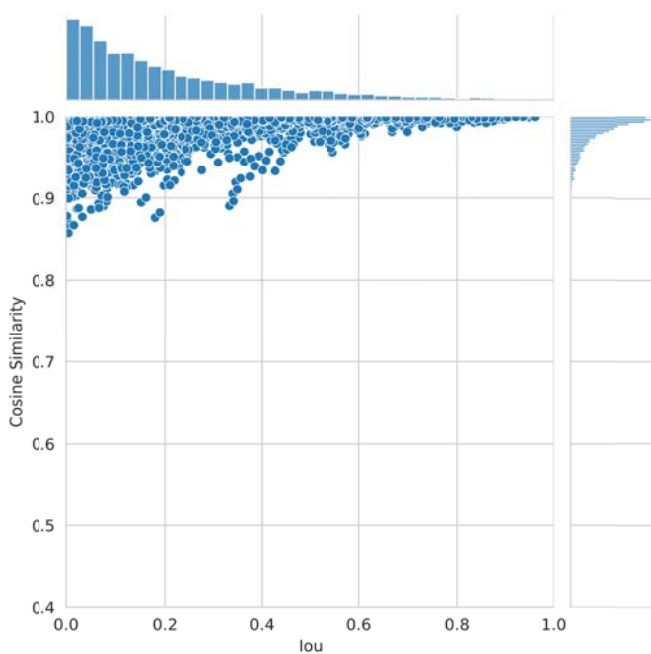

**Figure 3.** Statistics of IoU distance and cosine similarity among queries.

Homogeneous queries can lead to duplicate prediction problems. We explain this with an intuitive example, assuming that the two queries are very close to each other at the initialization time; for the sake of generality, we adopt the binary cross-entropy loss as the classification loss and the cross-entropy loss L for a single query can be expressed as follows:

$$L = -\left[ y \cdot \log\left( p \right) + \left( 1 - y \right) \cdot \log\left( 1 - p \right) \right] \tag{1}$$

where $y$ denotes the one-hot classification label corresponding to the GT and $p$ denotes the confidence score.

In the case of two queries, if one query $q_1$ matches the target GT and the other query $q_2$ does not match the object, according to Equation (1), the total loss of the two queries is the following:

$$L_{1,2} = L_1 + L_2 = -\left[\log(p_1) + \log(1-p_2)\right] \qquad (2)$$

Assuming that the two queries are very close to each other at their initialization, due to the strong correlation between the content similarity and the relative IoU distance, in the extreme case where we assume that the two queries are the same, their confidence scores have $p_1 = p_2 = p$, in which case the total loss in Equation (2) is the following:

$$L_{1,2} = -\left[\log(p) + \log(1-p)\right] \qquad (3)$$

The gradient of the loss with respect to the confidence level is as follows:

$$\frac{\partial L}{\partial p} = \frac{1}{1-p} - \frac{1}{p} \qquad (4)$$

It can be seen that when the confidence score is $p > 0.5$, the gradient is positive, the loss decreases with the decrease in confidence, and the network reduces the confidence of both predictions; when $p < 0.5$, the loss decreases with the increase in confidence, so the network increases the confidence of both predictions during training. Eventually, when $p=0.5$, the network reaches the equilibrium point of backward gradient propagation, at which there is no updated gradient for both queries, and the local optimum point of the loss function is reached. That is, in this case, the network prefers to keep the two low-confidence predictions instead of a de-duplication strategy that increases the confidence of one prediction to eliminate the other. As queries become dense and similar, more similar queries may correspond to a GT, generating duplicate predictions.

Homogeneous queries also bring unstable backpropagation during the training, which reduces the learning efficiency of the encoder. As in Figure 4, we assume that there are two queries that are close to the target GT, and in the two-part graph matching algorithm, query $q_1$, which matches GT produces a gradient update matrix $M_1$ that increases the confidence score of $q_1$, while query $q_2$ that matches no object produces a gradient update matrix $M_2$ that decreases the confidence score of $q_2$.

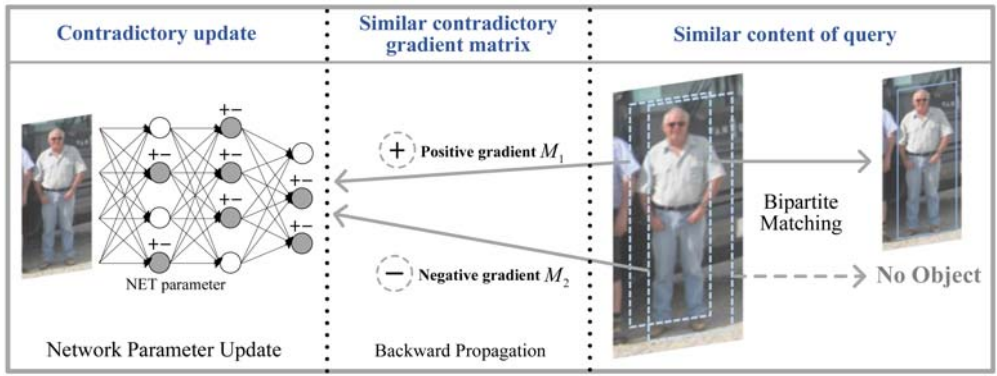

(a)

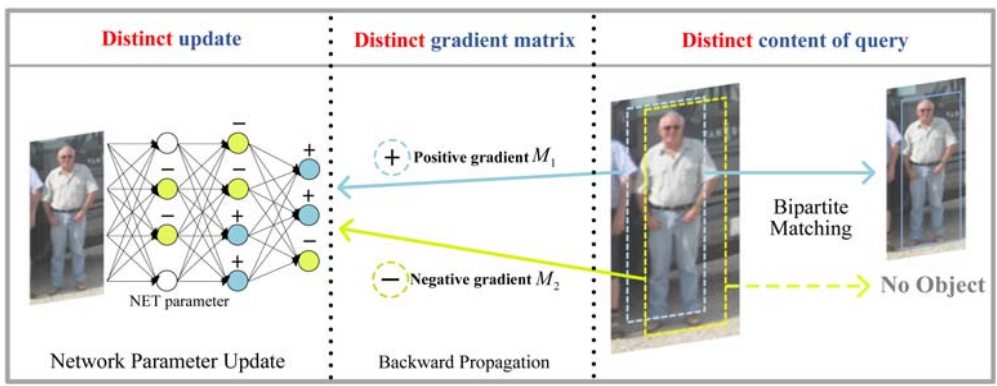

(b)

**Figure 4.** Network updates for different queries in training, and different colored boxes represent differences in the encoded content of the queries. (**a**) Detection methods based on homogeneous queries; (**b**) our proposed detection method based on de-homogenized queries.

As shown in Figure 4a, when the contents of two queries $q_1$ and $q_2$ are very similar, the resulting updated matrix $M_1$ and $M_2$ has similar absolute values but opposite sign directions. In the extreme case, when $q_1 = q_2$, the updated matrices $M_1$ and $M_2$ have the same absolute value but an opposite sign, that is $M_1 = -M_2$. Such contradictory updates make it difficult for the network's encoder to learn efficiently in dense scenarios, and the repeated predictions are difficult to be penalized by the back-propagation gradient in training.

As shown in Figure 4b, after adding differential coding to $q_1$ and $q_2$, even when their positions are almost coincident, the gradient update matrices $M_1$ and $M_2$ generated by the backpropagation of different symbol directions are significantly different from each other in absolute value. Such differentiated updates allow the encoder to learn more efficiently, and this model can learn the de-duplication strategy better through the differentiated information.

## 3. Our Method

The general framework proposed in this paper is shown in Figure 5, where the image is first passed through a feature extractor (including a backbone network and a stacked encoder) to obtain multi-scale feature maps, and then a fixed number of queries are initialized using our proposed GIoU-aware query selector, which is input to the subsequent decoder, and output location predictions through the first layer of the decoder with auxiliary prediction heads aligned to the encoder. Positional prediction is output through the first layer of the decoder aligned with auxiliary prediction heads. Then, the De-Homo Coding Generator (DCG) generates differentiated codes, which are added to the original query to form a differentiated query, which is then passed to the subsequent decoder to obtain the final confidence prediction.

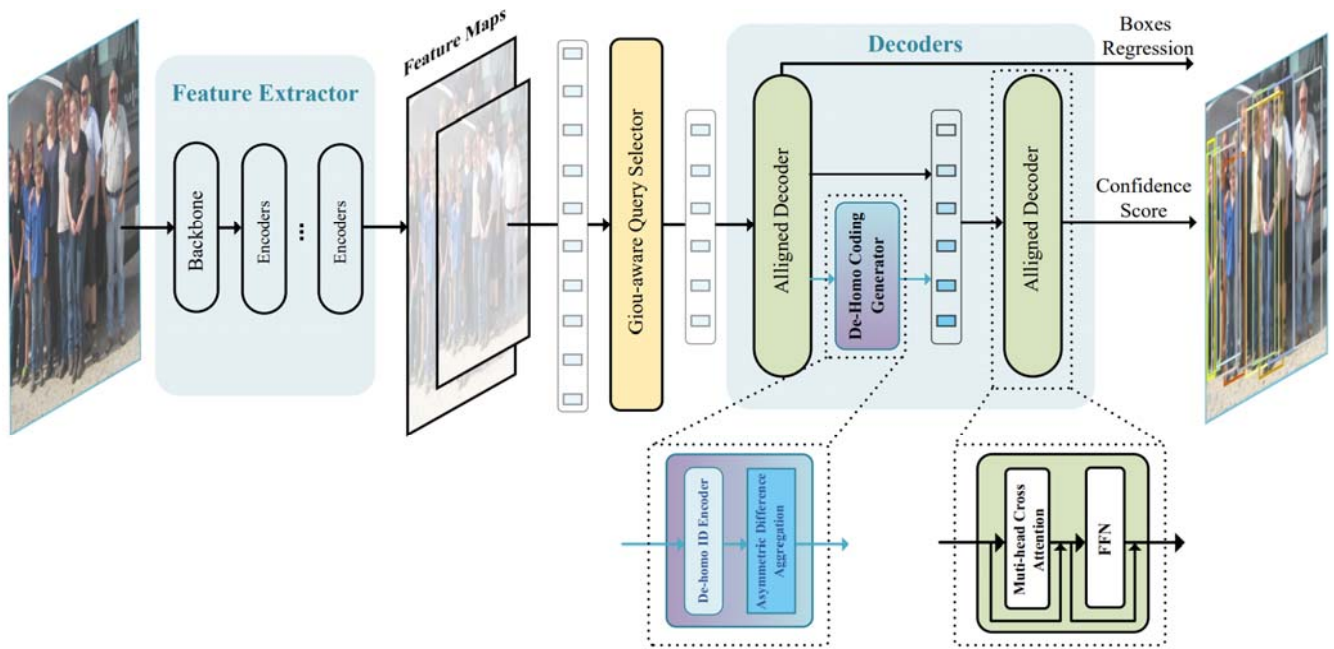

**Figure 5.** Overall framework of our proposed detector.

### 3.1. De-Homo Coding Generator

As mentioned above, homogeneous queries make it difficult for the network to learn effective de-duplication strategies. In this paper, we design a DCG (De-Homo Coding Generator) module to generate differentiated coded information to add to the query, which enables the network to learn the de-duplication ability through the differentiated information.

The DCG module, as illustrated in Figure 6, operates in two stages. In the first stage, a De-Homo ID Encoder generates a unique De-Homo ID for each query to learn the distinctions among them. The De-Homo ID for the $i$ th query, $e_i^{id}$, is calculated as follows:

$$e_i^{id} = LN\big(\mathcal{H}(q_i)\big) \tag{5}$$

where $\mathcal{H}(\cdot)$ encodes the query using two fully connected layers and an activation function. $LN$ represents layer normalization, which normalizes the encoded De-Homo ID for subsequent difference computation.

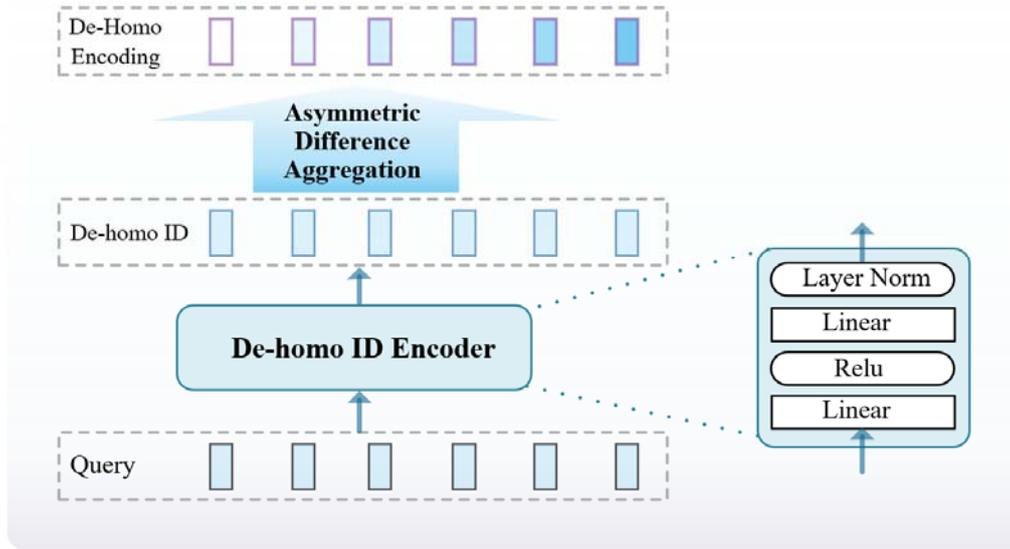

**Figure 6.** Structure of the DCG (De-Homo Coding Generator) module.

In the second stage, each query integrates the difference information from the De-Homo IDs of surrounding queries through an asymmetric difference fusion mechanism, producing differentiation encoding for de-duplication. This fusion mechanism also facilitates information exchange between queries, serving as an alternative to self-attention among them. For the $i$ th query, the differentiation encoding $q_i^{DE}$ is given as follows:

$$q_i^{DE} = ADA\big(e^{id}\big) = Maxpooling\left(\left\{\big(e_i^{id} - e_j^{id}\big)\cdot \mathbb{I}\big(c_j > c_i\big)\mid IoU(b_j, b_i) < 0.5, c_j > C_{low}\right\}\right) \tag{6}$$

where $c_i$ and $b_i$ represent the confidence and positional predictions of the $i$ th query, respectively. $\mathbb{I}$ is an indicator function that is 1 when $c_j > c_i$ and is 0 otherwise. The minimum threshold for confidence attention $C_{low}$ is set to avoid the computational complexity increase due to a large number of low-confidence predictions.

The asymmetric relationship based on confidence can further reinforce differentiated information, allowing each query to integrate the information of predictions with higher confidence scores from their surrounding. In traditional self-attention mechanisms among queries, where $q_j = W^\top \sum_{i=1}^{N} k_i W q_i$, the focus is only on information that improves recall without considering the differentiation from surrounding predictions to avoid redundancy. Our paper introduces the asymmetric difference aggregation (ADA) mechanism, which is a function that processes the difference of encoded content between queries $q_i - q_j$ to encode differentiated information.

The de-homogenized query is obtained by adding $q_i^{DE}$ to the original query, where $q_i^{De-Homo} = q_i + ffn(q_i^{DE})$. Here, $ffn$ represents a fully connected feedforward network consisting of two linear layers and an activation layer. $q_i^{De-Homo}$ is then fed into subsequent decoders to generate confidence predictions, addressing the issue of the network's difficulty in learning deduplication capabilities due to homogenized queries.

### 3.2. GIoU-Aware Query Selector

For the past two-stage query-based detectors [9,10,13,18], most algorithms use the Top-K algorithm with confidence scores to filter the queries from the encoder's predictions for initialization. However, the quality of the query's initialization is determined by the combination of the position and the confidence level, which results in the initialized query with a high IoU but a low confidence level being filtered out or a prediction with a low IoU but a high confidence level being selected, leading to the degradation of the quality of the query's initialization.

To solve this problem, we used a combined quality score that considered both confidence and location for query initialization. At the same time, the GIoU [19] can better reflect the overlap between prediction boxes based on IoU, so we use GIoU to indicate the quality of location prediction. We finally used Top-K based on the combined scores of GIoU and the Classification score to filter for query initialization. Meanwhile, we still used bipartite graph matching for a one-to-one mapping between predicted and real boxes.

We propose supervising the training of the encoder's output using GIoU-aware's combined predictive quality loss, which is as follows:

$$\mathcal{L}(\hat{y}, y) = \mathcal{L}_{box}(\hat{b}, b) + \mathcal{FL}_{giou-cls}(\hat{c}, c, Giou) \qquad (7)$$

where $\hat{y}$ and $y$ denote the prediction and GT, $\hat{y}$ consists of the position prediction $\hat{b}$ and the category prediction $\hat{c}$, $y$ also corresponds to the position label $b$ and the category label $c$ of the GT. We incorporate the idea of focal loss [20] to consider the loss that can focus more on hard samples, and our proposed GIoU-aware loss is as follows:

$$\mathcal{FL}_{giou-cls}(\hat{c}, c, Giou) = \begin{cases} -Giou \cdot \omega \cdot \log(\hat{p}) & \text{if } c = 1 \\ -\omega \log(1-\hat{p}) & \text{if } c = 0. \end{cases} \qquad (8)$$

where $\omega$ is modulating weight, which is similar to the idea of focal loss, and is used to determine the attention degree of difficult and easy samples, where the expression is as follows:

$$\omega = (\hat{p} \cdot Giou + (1-\hat{p})(1-Giou))^{\gamma} \qquad (9)$$

where $\gamma > 0$ is an adjustable factor. For simple samples where the predicted value is already very close to the true value, $\omega$ tends to zero, and for hard-to-predict samples, $\omega$ is given a higher weight.

In our test, we used the joint score Top-k for query initialization, which considers both location and confidence, to avoid the filtering out of low confidence and high IoU predictions caused by only confident scores so that we could achieve higher quality query initialization and higher accuracy with fewer queries. As shown in Figure 7, the predicted scores after using GQS have a better correlation with the IoU, resulting in higher IoU predictions when using the score Top-k algorithm, which corresponds to a better initialization quality. In the decoder loss design, in order to maintain the convergence of the model loss function, we still used the original discrete-label classification loss for supervised learning to obtain the final predictions.

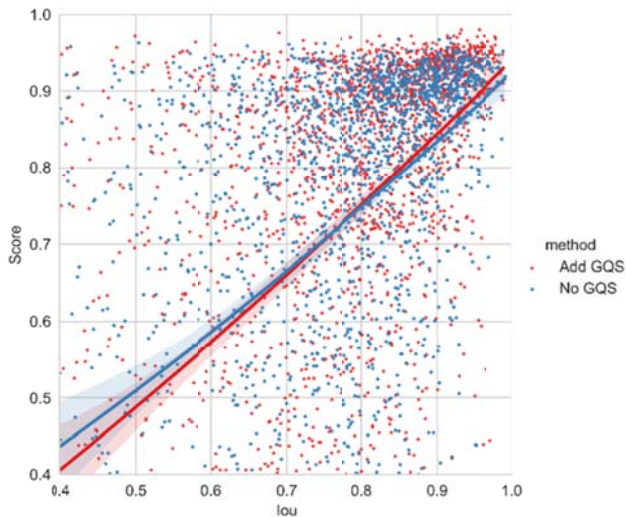

**Figure 7.** Statistics of score and IoU scores in initialized queries. The red line is the regression curve after the use of the GQS, and the blue line is without it.

### 3.3. Aligned Decoder

Many previous works attempted to reduce the differences between encoding and decoding methods by aligning the encoder and decoder so the model could integrate the information more efficiently in the cross-attention mecha-nism. Since the difference fusion mechanism in our proposed DEG was also available to perform information fusion among queries, we eliminated the self-attention module between the queries and used a decoder aligned with the encoder. While maintaining the original method's accuracy, we reduced the number of parameters by about 8%. With the addition of differential coding, both the encoder and decoder results improved because of the more stable one-to-one mapping relationship obtained in training. We also found that using only the first few layers of the trained decoder in the prediction stage reduced the number of parameters and maintained a relatively high accuracy.

## 4. Experiment and Analysis

We focus on the experimental validation of our approach through a benchmark test dataset, CrowdHuman [21], and compare it with SOTA's anchor-based detection method and query-based end-to-end detection method. We also performed ablation experiments for our proposed components and tested their robustness in scenarios with different densities.

### 4.1. Experiment Dataset

The CrowdHuman dataset contains 15k training images and 4.4k validation images, with an average of about 23 mutually occluded targets per image. We used the AP, $MR^{-2}$ and JI as metrics as follows:

- Average precision (AP): Expressed by the area enclosed by the precision–recall curve and coordinates. In object detection, AP is often used to reflect precision and recall, and a larger AP indicates better detection performance.
- $MR^{-2}$: The average missing detection rate is calculated on the logarithmic scale of the false detection rate for each image. This metric is often used in pedestrian detection because it reflects the false and missed detection, and a smaller $MR^{-2}$ indicates better detection performance.
- Jaccard index (JI): The index mainly evaluates the degree of overlap between the predicted set and the GTs. It reflects the overall distribution similarity between the prediction boxes set and the actual GTs, and a higher JI indicates better detection performance.

### 4.2. Experiment Details

We used the standard ResNet-50 [22], pre-trained on ImageNet [23], as the backbone for deformable DETR [18] and ran 50 epochs for training. We trained our model with the AdamW optimizer, where the momentum was set to 0.9 and the weight decay to 0.0001. The model's learning rate was 0.0002, and the learning rate of the backbone network was 0.00002. The batch size was 8, the number of attention headers was set to 8, and 4 RTX 3090 GPUs were used for training.

The end-to-end detector DETR [7], deformable DETR [18], Iter-E2Edet [15], etc., employed in our experiments used the default 6-layer encoder and 6-layer decoder; our method was implemented using only the 6-layer encoder and 3-layer decoder, and other hyper-parameter settings are kept the same as deformable DETR.

### 4.3. Comparative Experiments with Other Advanced Detectors

Table 1 shows a comparison of the results of our method on the CrowdHuman validation set with those of several SOTA methods, including anchor-based detectors [3,4,6,16,20,24–26] and query-based detectors [7,8,15,18,27,28]. It can be seen that the traditional anchor-based detection methods perform poorly in dense and high occlusion scenarios because they need to be de-duplicated by greedy algorithms such as NMS. In contrast, end-to-end detection methods learn the de-weighting inside the network and perform better overall. However, query-based methods usually need to continuously increase the number of queries to adapt to performance in dense scenarios, and most of the methods need to fine-tune the results hierarchically by stacking the decoder structure to improve the detection effect.

**Table 1.** Comparative experimental results on the CrowdHuman validation set.

| Method | #Queries | AP ↑ | $MR^{-2}$ ↓ | JI ↑ | Params |
|---|---|---|---|---|---|
| *Anchor-based detectors* | | | | | |
| RetinaNet [20] | - | 85.3 | 55.1 | 73.7 | |
| ATSS [24] | - | 87.0 | 55.1 | 75.9 | |
| ATSS [24] + MIP [16] | - | 88.7 | 51.6 | 77.0 | |
| Faster R-CNN [3] | - | 85.0 | 50.4 | - | |

10 of 17

| | | AP ↑ | MR⁻² ↓ | JI ↑ | Params(M) |
|---|---|---|---|---|---|
| Cascade R-CNN [4] | | 86.0 | 44.1 | - | |
| FPN [25] + Adaptive-NMS [6] | - | 84.7 | 47.7 | - | |
| FPN [25] + Soft-NMS [5] | - | 88.2 | 42.9 | 79.8 | |
| FPN [25] + MIP [16] | - | 90.7 | 41.4 | 82.3 | |
| PBM [26] | - | 89.3 | 43.3 | - | |
| *Query-based detectors* | | | | | |
| DETR [7] | 100 | 75.9 | 73.2 | 74.4 | |
| PED [27] | 1000 | 91.6 | 43.7 | 83.3 | |
| Sparse-RCNN [8] | 500 | 90.7 | 44.7 | 81.4 | |
| D-DETR (one-stage) [18] | 500 | 89.1 | 50.0 | | 37.7 M |
| | 1000 | 91.3 | 43.8 | 83.3 | 37.7 M |
| D-DETR (two-stage) [18] | 500 | 92.6 | 43.1 | 82.9 | 37.7 M |
| | 1000 | 92.8 | 43.2 | 83.0 | 37.7 M |
| UniHCP (direct eval) [28] | | 90.0 | 46.6 | 82.2 | 109.1 M |
| UniHCP (finetune) [28] | | 92.5 | 41.6 | **85.8** | 109.1 M |
| Iter-E2EDet [15] | 500 | 91.2 | 42.6 | 84.0 | 38.0 M |
| | 1000 | 92.1 | 41.5 | 84.0 | 38.0 M |
| Ours(6-3)* | 500 | **93.6** | **39.2** | 84.3 | 34.6 M |
| Ours(6-3(2))* | 500 | 93.5 | 39.3 | 84.1 | **33.7 M** |

* The numbers in parentheses after the method name indicate the number of layers of the encoder and decoder used for training and testing. X-Y (Z) indicates the X-layer encoder and Y-layer decoder for training and the Z-layer decoder for testing. If no special instruction exists, other methods are 6-6 (6) by default.

Comparing these methods, using fewer queries, our method achieved higher accuracy and lower miss detection rates. Our method reduced about 3.9% and improved AP by 1% compared to the benchmark method of two-stage deformable DETR using six decoders, while our model reduced the number of parameters by about 8% in the parameter scale. UniHCP [28] was first trained across tasks on 33 datasets with about 109.1 million parameters, which were then fine-tuned on the Crowdhuman dataset. Our method still outperformed them in AP and MR⁻² metrics, using only about 32% of the parameters.

### 4.4. Ablation Experiment

To test the effectiveness of the different components in our proposed method, we performed ablation experiments on Crowdhuman. As in Table 2, in the first row of the table, we used six decoders of deformable-DETR (ResNet50 as the backbone network) as the benchmark method for comparison, and three decoders were used by default in our method. Our proposed De-Homo Coding Generator (DCG) significantly improved the detector's performance, with the leakage rate MR⁻² reduced from the original 43.2% to 40.2%. After using the GIoU-aware query selector, GQS, the miss prediction was further reduced because of the improved initialization efficiency of the query.

**Table 2.** Ablation experiments on CrowdHuman validation set.

| DCG | AD | GQS | AP ↑ | MR⁻² ↓ | JI ↑ | Params(M) |
|---|---|---|---|---|---|---|
| | | | 92.8 | 43.2 | 83.0 | 37.7 |
| ✓ | | | 93.3 | 40.2 | 83.8 | 35.4 |
| ✓ | ✓ | | 93.5 | 40.0 | 84.0 | 34.6 |
| ✓ | ✓ | ✓ | **93.6** | **39.2** | **84.3** | **34.6** |

* DCG—De-Homo Coding Generator. AD—aligned decoder. GQS—GIoU query selector.

At the same time, our method did not need a redundant layer fine-tuning structure and obtained high accuracy at the early stage of the decoder, as shown in Figures 8 and 9. We tested the AP and MR$^{-2}$ of the prediction results in our method and the benchmark method deformable DETR for different stages of the decoder. In the second decoder block, our model achieved an accuracy of 93.5% for AP and 39.3% for MR$^{-2}$, which basically guaranteed accuracy while reducing the number of parameters. Comparing the outputs of the encoder and the first decoder, our method significantly reduced the information gap between the decoder and the encoder.

It can also be seen that when the encoder's structure is exactly the same, its performance improves significantly due to the distinctive gradient update brought about by the differentiated query.

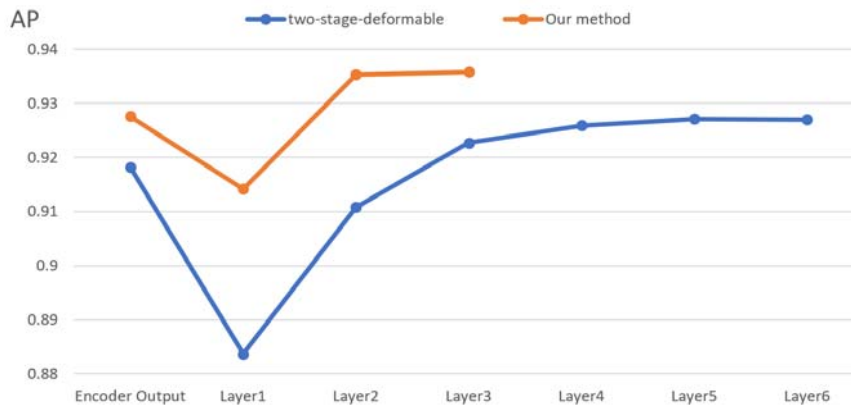

**Figure 8.** AP of detection results for each stage of the decoder.

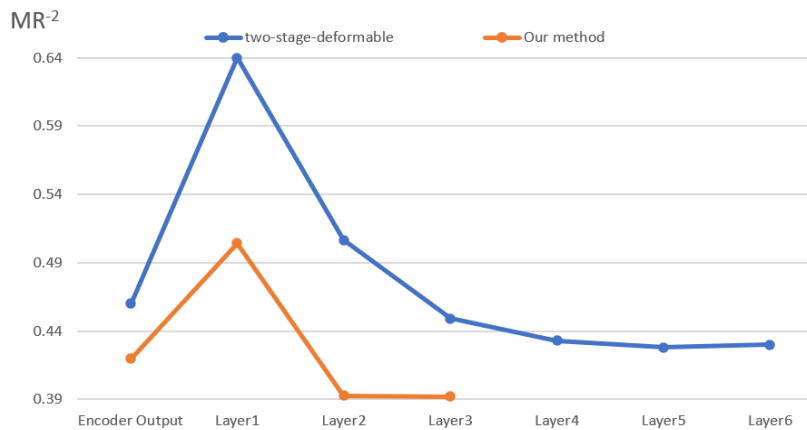

**Figure 9.** MR$^{-2}$ of detection results for each stage of the decoder.

### 4.5. Hyperparameter Analysis

Table 3 compares the model's performance on the CrowdHuman validation set when different decoders are assigned before and after the differential coding generator DCG. The results are better when more decoders are used after differential coding.

**Table 3.** Performance of the model when different numbers of decoders are used before and after the DCG.

| Decoders (before) | Decoders (after) | AP ↑ | MR$^{-2}$ ↓ | JI ↑ | Params (M) |
|---|---|---|---|---|---|
| 1 | 1 | 93.3 | 40.1 | 84.1 | 33.7 |
| 2 | 1 | 93.4 | 40.3 | 84.2 | 34.4 |
| 1 | 2 | 93.6 | 39.2 | 84.3 | 34.6 |
| 1 | 2(1) [1] | 93.5 | 39.3 | 84.1 | 33.7 |

[1] Here, 2(1) means that 2 decoders are used for training, but only 1 decoder is used for testing.

At the same time, we compared the impact of using different queries, as shown in Table 4; our method improved the benchmark method when using various queries. Figure 10 shows the performance trend of the model with different numbers of queries, and it is clear that our method can perform better with fewer queries. Performance degradation tends to be slower when the number of queries becomes smaller.

**Table 4.** Performance of the model with different numbers of queries in the Crowdhuman validation set.

| #Queries | Our Method | | Deformable DETR | |
|---|---|---|---|---|
| | **AP ↑** | **MR⁻² ↓** | **AP ↑** | **MR⁻² ↓** |
| 100 | 88.93 | 39.87 | 87.59 | 47.56 |
| 200 | 92.53 | 39.24 | 91.40 | 44.35 |
| 300 | 93.26 | 39.21 | 92.23 | 43.44 |
| 500 | 93.58 | 39.20 | 92.61 | 43.10 |
| 1000 | 93.75 | 39.20 | 92.77 | 43.19 |
| 2000 | 93.77 | 39.21 | 92.79 | 43.25 |

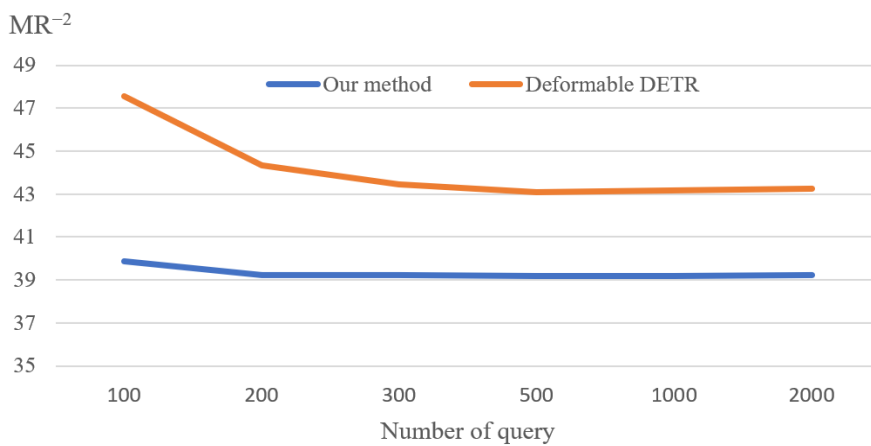

**Figure 10.** Comparison of model performance when using different numbers of queries. Comparative analysis of differentiated queries.

To compare the effect of our method on query differentiation, as shown in Figure 11, we computed the IOU distances and the cosine similarities between the queries with positive predictive confidence in the deformable DETR and our method, respectively. Our method can significantly reduce query homogeneity and break the strong correlation between query location and the encoded content.

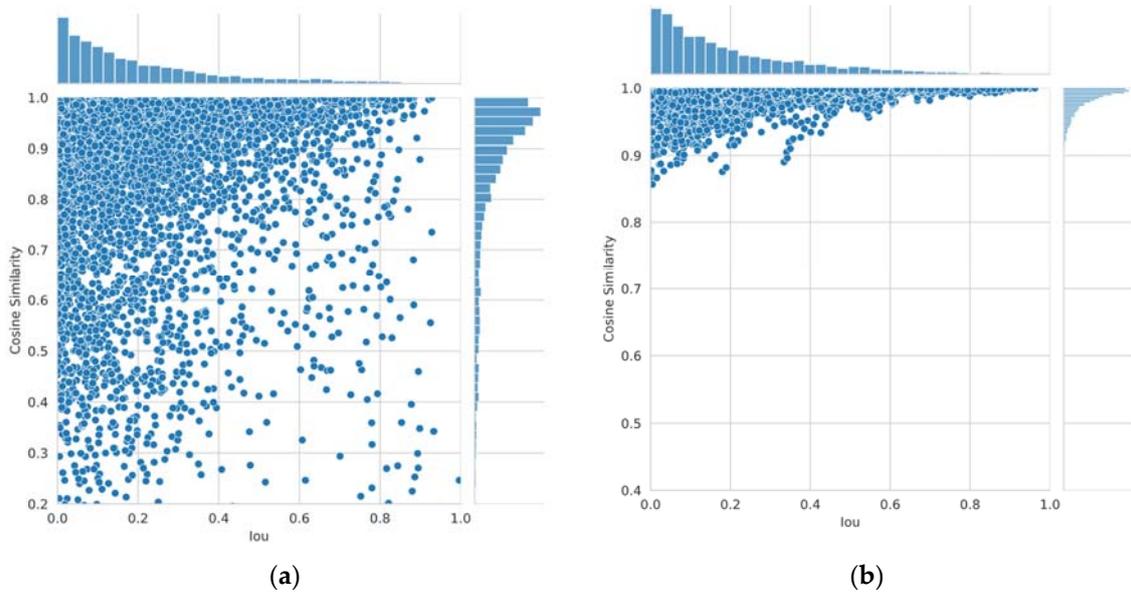

(**a**)                                                   (**b**)

**Figure 11.** Cosine similarity of query at different IoU distances (**a**) before de-homogenization, (**b**) after de-homogenization.

*4.6. Analysis of Detection Results*

We analyzed the detection results of our method and the benchmark method two-stage deformable-DETR in detail. We also made false positive (FP) and true positive (TP) statistics at different confidence scores in the detection results of the CrowdHuman validation set. The statistical results are shown in Figure 12, with the matching IOU threshold set to 0.8. Our method significantly improved the TP at each confidence level while largely suppressing the FP of repeated predictions.

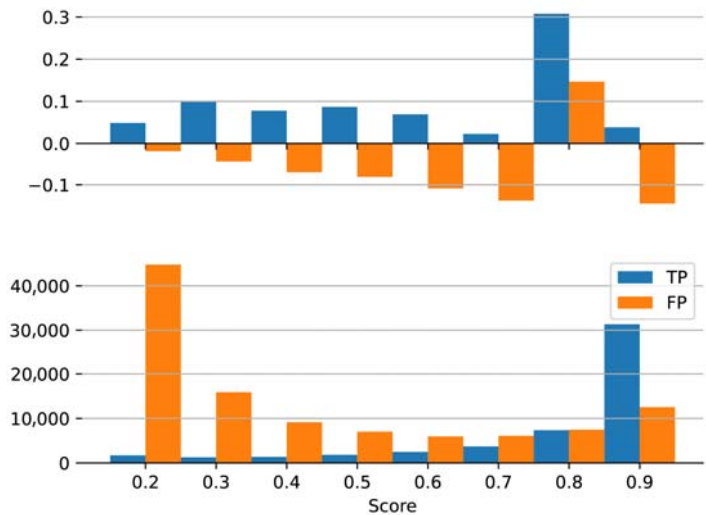

**Figure 12.** Comparing the relative improvement of our detection results in different confidence scores.

To verify the generalization performance and robustness of our method in scenarios with different densities, we statistically measured the changes in TP and FP between our method and the benchmark method in scenarios with different densities. As shown in Figure 13; when we set the matching IOU threshold to 0.5, our method significantly increased the TP and inhibited the repeated prediction of FP in various scenarios with different densities. When the IOU threshold was set to 0.8, as shown in Figure 14, the improvement of our method over the benchmark method was even more significant. Our method can regress the target location more accurately and simultaneously reduce the FP of repeated prediction and false detection.

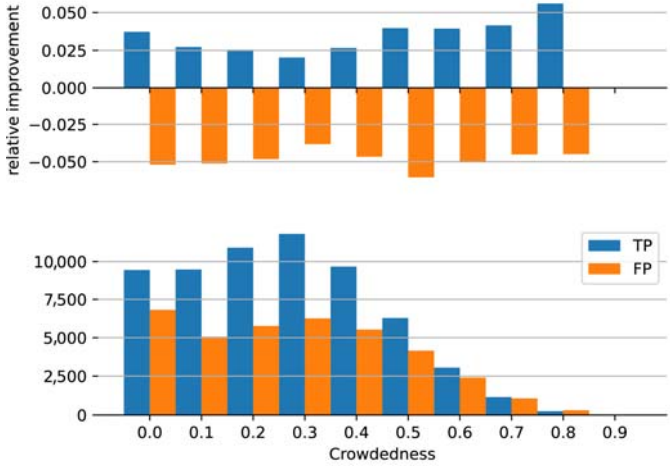

**Figure 13.** The relative improvement of our method over deformable DETR in different dense scenarios (the matching IoU threshold is 0.5).

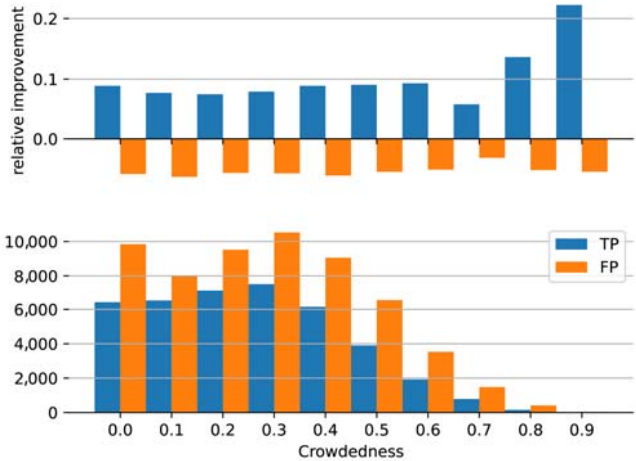

**Figure 14.** The relative improvement of our method over deformable DETR in different dense scenarios (the matching IoU threshold is 0.8).

*4.7. Comparison of Actual Test Result Images*

As shown in Figure 15, we compared the detection effect of two-stage deformable DETR with that of our method and found that our method still performed well in dense and heavily occluded scenes.

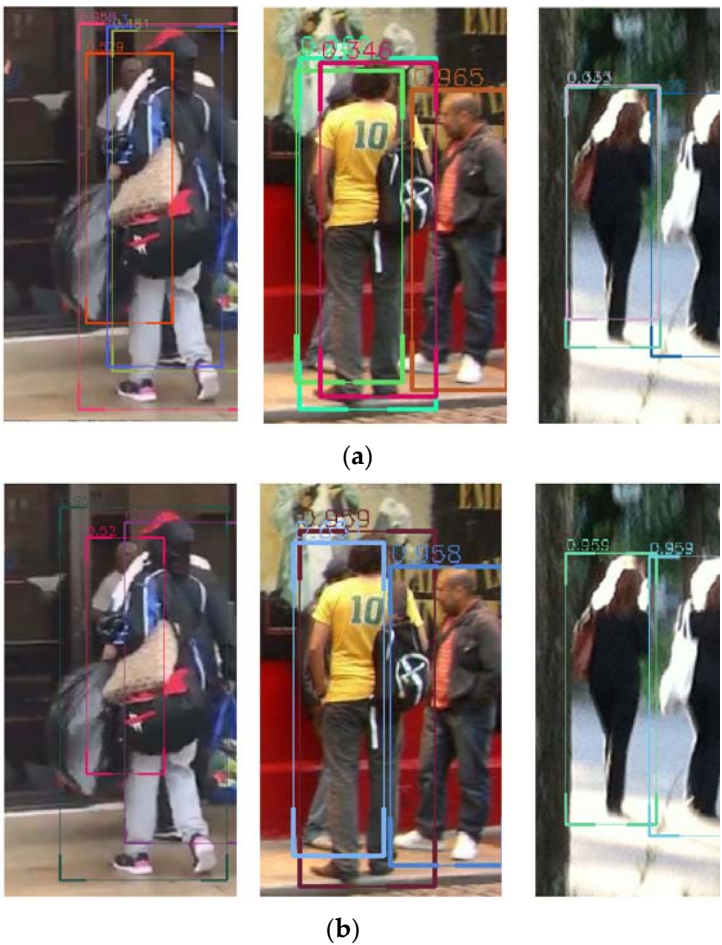

(**a**)

(**b**)

**Figure 15.** Comparison of the actual detection results for (**a**) the two-stage deformable DETR (**b**) and our method.

## 5. Discussion

In this paper, we propose a differentiated query strategy, which significantly increases the de-duplication capability of the query-based detection model and improves the learning efficiency of the encoder in dense scenarios; at the same time, we optimize the initialization of the query and the structure of the decoder, which reduces the number of parameters in the model while improving the accuracy of the end-to-end detector. Compared with the current SOTA detection methods, this method achieves higher accuracy while keeping the number of parameters lower, and its robustness in different dense scenes is experimentally verified. Nevertheless, our proposed DCG (De-Homo Coding Generator) module leads to a higher time complexity in the inference phase, especially in dense scenarios, because of the computation of IoU among dense queries. In addition, to avoid excessive time complexity, we limited the computational scope of the De-Homo Coding and the depth of the network for the coding methods. Our approach still has potential for improvement in complex and dense scenarios. Also, the initialization of the query using encoder features incurs an extra high computational complexity in inference. Future work will further optimize the model based on this to enhance its robustness in more complex scenarios and further optimize the computational complexity of the model.

**Author Contributions:** Conceptualization, Y.H. and G.Y.; methodology, Y.H. and G.Y.; software, Y.H. and C.M.; validation, Y.H. and C.M.; data curation, H.W.; writing—original draft preparation, Y.H. and G.Y.; writing—review and editing, H.Z. and G.Y.; visualization, H.W.; supervision, H.Z. and G.Y.; funding acquisition, H.Z. All authors have read and agreed to the published version of the manuscript.

**Funding:** This research was funded by the Natural Science Foundation of China (grant no. 62162065, 62061049, 12263008), the Department of Science and Technology of Yunnan Province–Yunnan University Joint Special Project for Double-Class Construction (grant no. 202201BF070001-005).

**Data Availability Statement:** The CrowdHuman dataset is available at http://www.crowdhuman.org(accessed on 1 June May 2023)

**Conflicts of Interest:** The authors declare no conflicts of interest.